# Exploring Convolutional Neural Networks for Rice Grain Classification: An Explainable AI Approach

Muhammad Junaid Asif
*Artificial Intelligence Technology Centre (AITeC)*
National Centre for Physics (NCP)
Islamabad (44000), Pakistan
junaid.asif@ncp.edu.pk

Hamza Khan
*Faculty of IT and Computer Sciences*
University of Central Punjab (UCP)
Lahore (54590), Pakistan
L1S22MSDS0011@ucp.edu.pk

Rabia Tehseen
*Faculty of IT and Computer Sciences*
University of Central Punjab (UCP)
Lahore (54590), Pakistan
rabia.tehseen@ucp.edu.pk

Rana Fayyaz Ahmad
*Artificial Intelligence Tech Centre*
National Centre for Physics (NCP)
Islamabad (44000), Pakistan

Mujtaba Asad
*Department of Automation*
Shanghai JiaoTong University,
Shanghai (200240) China

Syed Tahir Hussain Rizvi
*Wenn ASA*
Sandnes, Norway
tahir.rizvi@wenn.no

Shazia Saqib
*School of Informatics and Robotics*
Institute of Arts and Culture (IAC)
Lahore, Pakistan

***Abstract*— Rice is a global staple food that plays a key role in terms of fostering global trade, economic developments and food. China, India, Pakistan, Thailand, Vietnam and Indonesia are some of the Asian countries that have been engaged in the production of various varieties of rice including short and long grains. These varieties are even further categorized into basmati, jasmine, kainat saila, ipsala, arborio, and so on, which suits all the different culinary styles as well as cultural backgrounds. In case of local and international trade, there is need to check and maintain the quality of rice grains to satisfy the customers and to maintain the reputation of a country. Quality check and classification using the manual are very tedious and time consuming. It is also very susceptible to errors. In this research paper, the automatic framework is a convolutional neural network (CNN) framework to classify rice varieties, i.e. basmati, jasmine, ipsala, arborio, and karacadag. The CNN model was trained and validated, with an impressive accuracy rate and an ideal area under each class Receiver Operating Characteristic (ROC) curve. The confusion matrix analysis indicated that the model was effective in differentiating the various rice varieties with few misclassifications. Also, the use of explainability methods including LIME (Local Interpretable Model-agnostic Explanations) and SHAP (SHapley Additive exPlanations) helped to understand the decision-making process of a model and provided some new insights about how particular features of rice grains impacted the results of classification. The interpretability fosters trust in the model predictions and increases its applicability in the real-world context. The results demonstrate the enormous potential of deep learning methods in agriculture, which opens the path to the development of automated classification systems.**

***Keywords*— *Crop Classification, Rice Grain Classification, Explainable AI, XAI, CNN, Deep Learning***

## I. Introduction

Agriculture includes the production of crops and farming animals to be used by people and thus giving economies a gain all around the world. It also encompasses animal husbandry, planting and harvesting [1]. The symbiotic relationship between agricultural practices and crop production guarantees a stable food supply, farming revenues, and the wellbeing of the world population [2]. Accurate and precise analysis of crops is key to this system, which allows optimization of yields, improvement of sustainability, and decision-making about the health of crops and resource distribution. Crop analysis enables the agricultural industry to use various technologies to come up with detailed analysis of crop health, classification, crop yields and effective resource allocation [3]. Satellite images, unmanned aerial vehicles, and remote sensing are examples of tools that complement crop production by collecting vast amounts of information about the state of the agricultural sector [4].

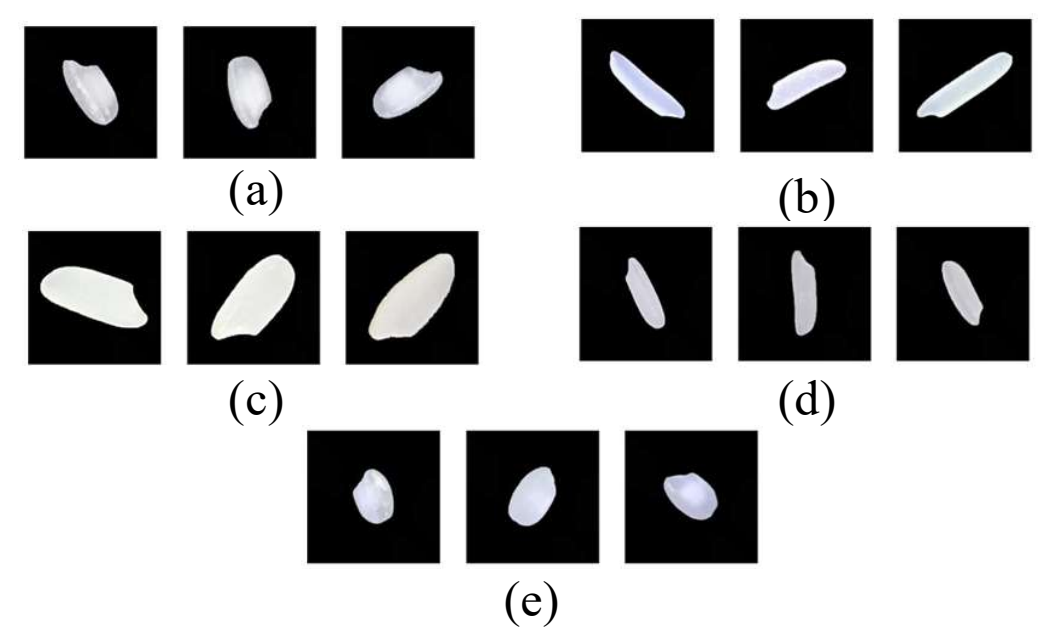

Fig. 1. Different Varieties of Rice Grains (a) Arborio Rice Grains, (b) Basmati Rice Grains, (c) Ipsala Rice Grains, (d) Jasmine Rice Grains, (e) Karacadag Rice Grains

This information can be analyzed using machine learning and artificial intelligence systems, and it can provide useful information on the detection of diseases, nutrient deficiencies, and the overall condition of crops [5]. The recent development of AI has encouraged agricultural scientists to explore the field of deep learning, with a focus on the application of AI to crop analysis. Many agricultural activities such as identifying problems, monitoring crop health [6], estimating yields [7] estimating prices [8] but also mapping yields, optimizing the use of pesticides [9], and fertilizers, and applying them, benefit greatly through the addition of AI [10]. A major phenomenon of crop analysis is crop classification, which involves classifying different types of crops. Proper categorization of crop type plays an essential role in appropriate agricultural management, as it helps to make predictions more effectively, make decisions based on data and practice a sustainable agriculture [11].

Rice, a major staple crop, is also grown in over 100 countries and is produced in massively in China and Southeast Asia. This is a highly important grain, which supports a population of about 3.5 billion people across the world, and it is the third most produced crop after maize and sugarcane [12]. Manual rice grains classification is time-consuming and tedious and requires careful inspection of grain size, morphology, and color. This is not economical and efficient when dealing with large operations. A practical alternative is automated classification systems, which combine computer vision (CV) and deep learning (DL), and in a manner akin to a large-scale analysis of large datasets, the systems can classify and identify rice varieties based on a predefined attribute. It leads to faster and more precise classification, cost and time savings, reduced human error, and better identification consistency of the rice type [13] . In this paper, an automatic system of categorizing different types of rice ***(as shown in Fig. 1***) that is, Arborio, Basmati, Jasmine, Ipsala and Karacadag, will be proposed using a combination of parameters such as grain size, type of grain and other pertinent features.

Convolutional neural networks (CNNs) are among the most effective deep learning architectures, especially useful in the processing and analysis of visual impingements. CNNs produce hierarchical features representations of raw images and thus yield high accuracy in classifying images such as rice grains [14]. In the agricultural sector, CNNs are particularly effective in identifying salient features and attributes of images of rice grains (including grain size, shape, colour, eccentricity, major axis length, and minor axis length) which can be used to distinguish between varieties. Significant CNN models, such as VGG16 [15], InceptionNet [16], and MobileNetV3Small [17] have been applied to various crop-analysis tasks with better accuracy. However, the lack of transparency of these models as opaque black-box models may hinder their usage in rice-grain classification problems, where interpretability and accountability are essential in making decisions which have a direct impact on product quality and supply-chain logistics.

In this paper, the proposed methodological framework combines CNNs with two different explainable AI (XAI) approaches: Local Interpretable Model-agnostic Explanations (LIME) [18], [19] and Shapley Additive explanations (SHAP) [18], [19] to obtain accurate classification of rice-grain varieties. To increase transparency we combine Explainable Artificial Intelligence (XAI) techniques, namely LIME (Local Interpretable Model-Agnostic Explanations) and SHAP (SHapley Additive exPlanations). These methods provide human-understandable explanations of the predictions made by the Convolutional Neural Network (CNN) by visualizing the impact of image features, e.g. grain lines and textures, on the model outputs. LIME focuses on super pixels effects of individual prediction, while SHAP offers a global view of feature relevance that is based on game-theoretic concepts. With the combination of the tools, we ensure that the model is accurate, interpretable, and reliable, which is critical to be used in the agricultural systems. The block diagram of our proposed system is ***(as shown in Fig. 2).***

The novelty of the current study is in the fact that the application of explainable AI methods (LIME and SHAP) has been incorporated in the integrated manner, to interpret and justify the predictions of the model at the granular level. In contrast to the previous studies which are interested only in the accuracy of the classification, the current study puts more emphasis on the transparency of the model and provides the visual descriptions of the learned features and the decision-making procedure concerning every type of rice grain. This two-explainability approach does not only support the reliability of the model but also provides human-understandable information on the significance of features, thus making the model more useful in terms of agricultural quality control and practical application. The main contributions of this paper will be as follows:

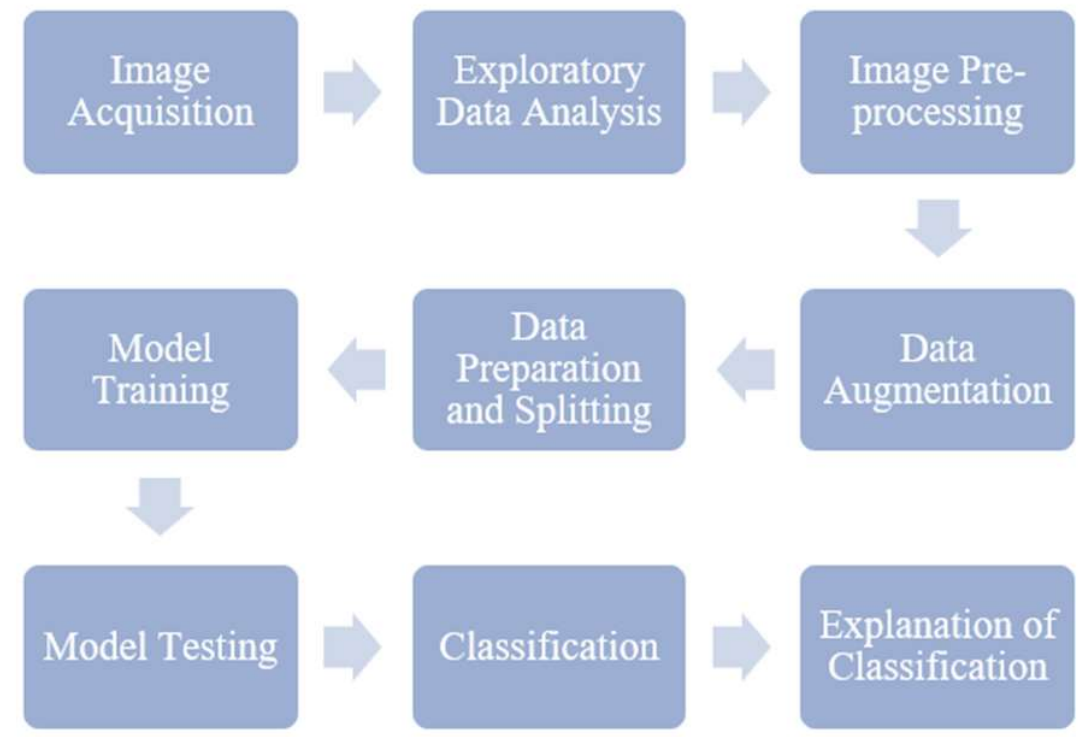

Fig. 2. Block Diagram of Proposed Rice Grain Classification System

- The major contribution is that explainable AI methods, namely SHAP and LIME, are integrated into CNNs to enhance the interpretability of rice grain classification models to make them more reliable and feasible to farmers, researchers, and industry professionals.
- A secondary contribution is the comparative examination of LIME and SHAP, which highlights the strong points of the two, namely the local interpretability of the former and the significance of the feature of the importance of the global aspects of the latter, to generate a more comprehensive understanding of the model explainability.

This paper is structured to provide a thorough understanding of the research. Section 2 covers related work and identifies research gaps. Section 3 outlines the proposed methodology and system flow. Section 4 details the materials, methods, and experimental setup. Section 5 presents experimental validation, including contrast experiments, ablation studies, and sensitivity analysis. Section 6 reports the results and offers a discussion in the context of existing literature.

## II. Literature Review

Recent developments in the field of artificial intelligence have seen a broad application of deep learning to various industrial applications, including that forecast passenger behaviour [20] to predictive maintenance of machinery [21] to video analysis [22], [23], [24]. The technology has also been successful in various classification problems, such as in cancer cell identification [25] and patient categorisation [26]. Deep learning approaches are gaining importance in the agrarian domain [14]. Crop analysis using deep learning has received much investigation, especially focusing on crop classification [27] and crop disease prediction [28]. Although there are a number of proposed and proven models, classifying rice grains is still a major challenge to farmers and quality analysts. This research paper introduces an automated system that can be used to classify rice varieties efficiently and accurately. The next section is a literature review on deep learning methods used in rice grain classification. A comparative analysis was conducted by Koklu et al. [29] between artificial neural networks (ANNs), deep neural networks (DNNs) when using feature datasets, and convolutional neural networks (CNNs) when using image datasets. Performance measures (F1-score, precision, accuracy, false positive rate (FPR), false negative rate (FNR), sensitivity, specificity) were used to assess it. The models were trained using image-based data consisting of 75,000 images of five varieties of rice grain namely, arborio, jasmine, basmati, ipsala and karacadag. The research achieved 100 mg to CNN, 99.95% to DNN, and 99.87% to ANN, which implies that the application will be applicable in the classification of various food grains.

In [30], the author developed a study in which seven machine-learning classifiers, including logistic regression (LR), decision tree (DT), support vector machine (SVM), random forest (RF), multilayer perceptron (MLP), Naïve Bayes (NB), and k-nearest neighbors (KNN) are used to classify rice varieties. The experimental outcomes were encouraging, and the random-forest classification was 99.85 % accurate and the decision-tree classification was 99.68% accurate. Another author, in [31], suggested that a multiclass SVM framework be used to classify three types of rice grain, basmati, ponni, and brown grains. The proposed research was tested with 90 test images and attained a classification of 92.22%. The study by Ramadhani et al. [32] focused on the optimization of SVM and KNN using genetic algorithms. They obtained 92.81% accuracy with SVM-GA and 88.31% with KNN-GA. The findings reveal that the combination of SVM and genetic algorithms is more effective than other algorithms, which may improve sustainable practices of cultivating rice and its productivity, in general [33]. Bejerano et al. [34] introduced a hybrid model that combines a deep convolutional neural network (DCNN) and support vector machine (SVM). The suggested framework was trained to be able to classify four types of rice, which are damaged, discolored, broken, and chalky. SVM was used to produce classification and DCNN produced morphological features. The grading of rice was tested and graded on a model with an accuracy of 98.33% and 98.75% on the training and validation data respectively.

A framework of classification of five types of rice grains using ResNet-50 was suggested by Rayudu et al. [35] which included basmati, jasmine, arborio, Uppsala, and karacadag. The model has been trained and tested on a small dataset consisting of 2,500 images (500 each) used. When compared to VGG16 and MobileNet, the ResNet-50 demonstrated higher results indicating the strong potential of deep learning to enhance the efficiency of rice categorization and, therefore, quality control, supply-chain management, and consumer confidence.

## III. Description & Methodology of Proposed System

The suggested methodology ***(as shown in Fig. 3)*** will help to improve the classification of rice grains with the help of a CNN that is backed up by explainable AI. Once data has been acquired, pre-processed, and splitted, the CNN is then trained in hierarchical features. LIME is used to get local interpretability whereas SHAP is used to get global insights. The last model is tested on a test set to provide correct unbiased predictions.

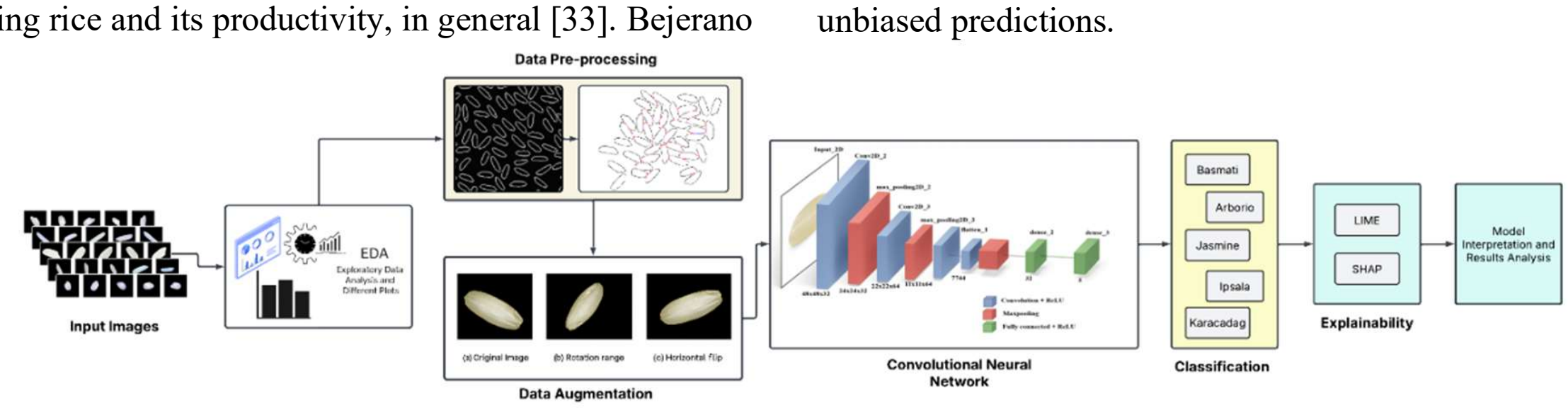

Fig. 3. Proposed Explainable Artificial Intelligence (XAI) Framework for Rice Grain Classification

### A. Image Acquisition and Preprocessing:

A publicly available dataset on Kaggle is the Rice Image Dataset [29], which is suggested to be used in the development of the system. The proposed dataset will have 75,000 images, that is, each rice grain variety will have 15,000 images.

***1. Edge Detection and Image Segmentation:*** To help with classification and analysis, the edge detector was used to emphasize the outline of the grains ***(as shown in Fig. 4).*** In this investigation, Canny edge detector that detects sharp variations in intensity and is useful in rice grain boundaries was used. Photos were image-to-grayscale, denoised through the Gaussian blur algorithm, and the Canny algorithm was applied ***(as shown in Fig. 5)***. This refines the grain morphology and yields better visualization which is a crucial preprocessing stage towards the correct recognition of rice varieties. After edge detection, image segmentation is used to outline separate regions, which can help isolate and analyze individual rice grains precisely. The effect of this step of the procedure is the enhancement of the classification accuracy due to the clear division of grains and background as well as grains themselves.

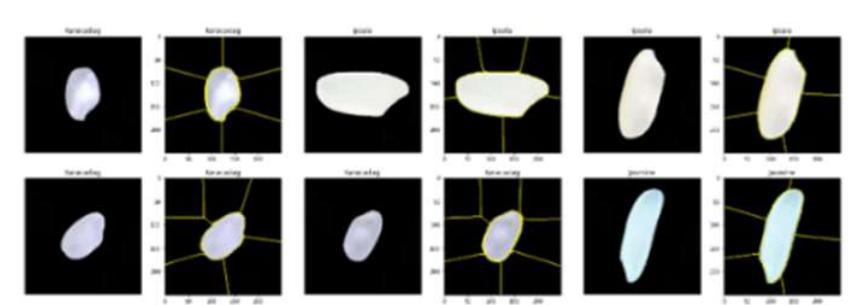

Fig. 4. Extraction of rice grain outlines using the Canny Edge Detection approach.

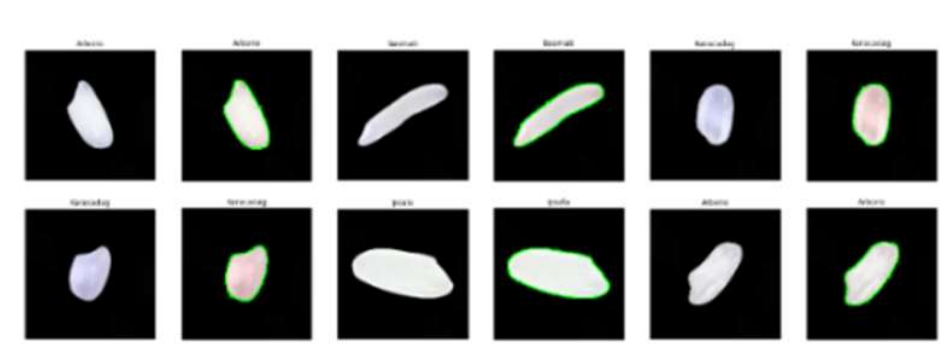

Fig. 5. Rice grains with detected edges, highlighting the contours and characteristics essential for classification and analysis.

***3. Data Normalization and Augmentation:***

The non-homogenous pixel values make the model convergence in training difficult. To overcome this, any segmented images undergo a normalization process that maps the pixel values in a scalable manner to a limited set of 0-1. All the images were augmented by two unique data augmentation processes: horizontal flipping and rotation to increase the data volume and variety. Images were flipped and rotated 90 and 180 and flipped at the probability values of -1 to +1.

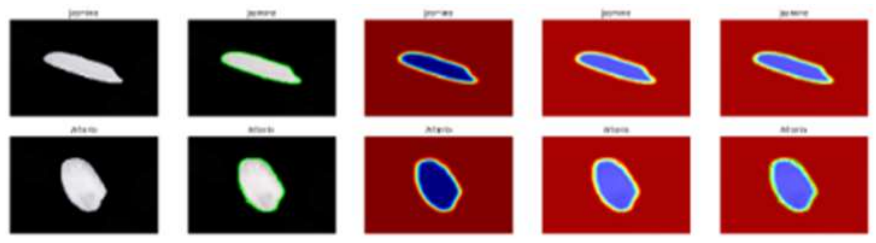

Fig. 6. Segmented rice grain images highlighting individual grain boundaries for precise analysis.

### *B. Proposed Model:*

In this paper, CNN ***(as shown in Fig. 7)*** is developed to categorize 50×50 RGB rice grain images with the assistance of Keras Sequential API. It has an opening Conv2D filter of 32 (3×3) with ReLU to obtain simple spatial elements and is followed by MaxPooling2D to be used to down sample. A second Conv2D and another pooling layer detect patterns on high level. The feature is flattened and then subjected to a 32-unit dense layer of ReLU to abstract features. The last SoftMax layer makes a prediction of five classes. The model is trained on the Adamax optimizer categorical cross-entropy and assessed based on accuracy.

TABLE 1. HYPERPARAMETERS DETAILS

| Hyperparameter | Value |
|---|---|
| **Batch Size** | 32 |
| **Image Size** | 50x50 |
| **No. of channels** | 03 (RGB) |
| **Optimizer** | ADAM |
| **Model Regularization** | L2 Regularization |

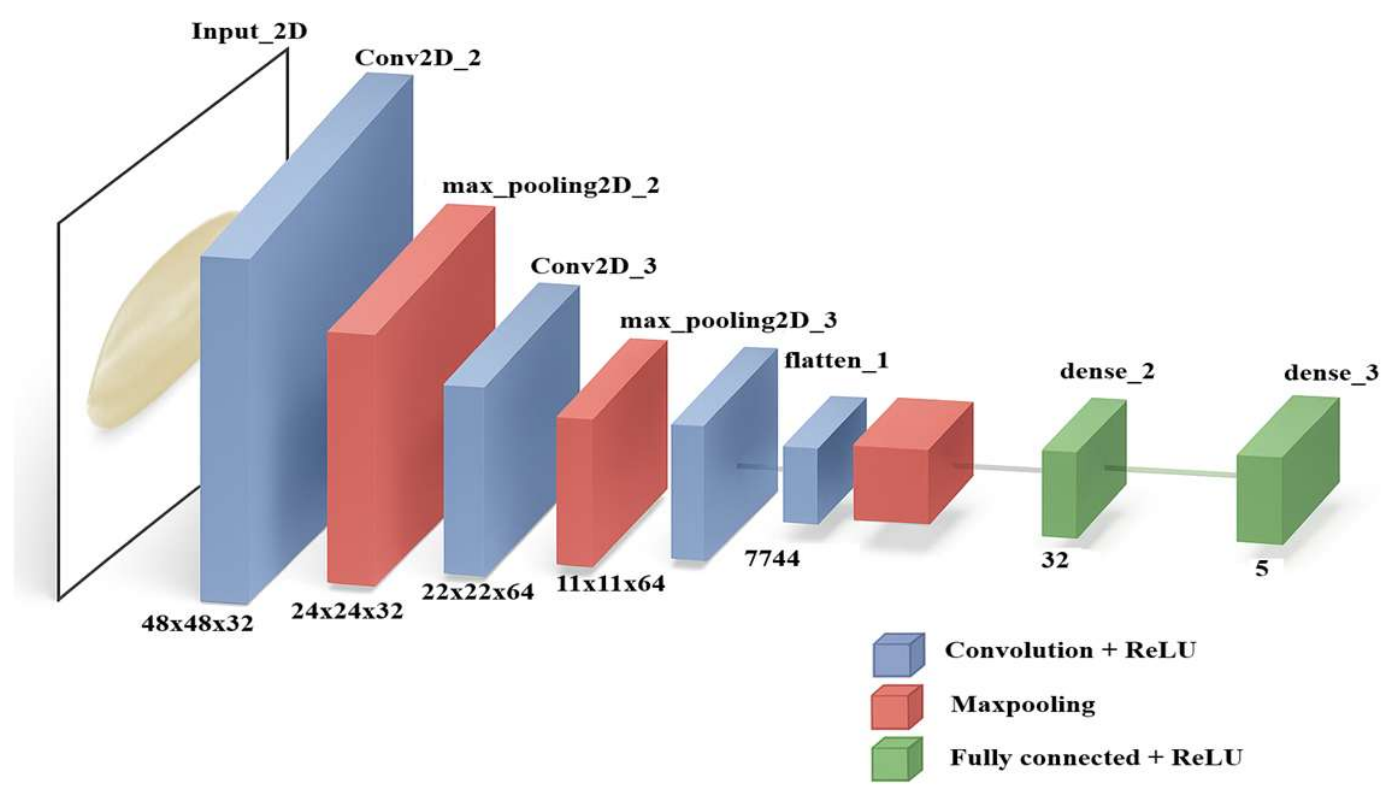

Fig. 7. Architecture of Proposed CNN Model

TABLE 2. ARCHITECTURE AND PARAMETER SUMMARY OF THE PROPOSED CNN MODEL FOR RICE GRAIN CLASSIFICATION.

| Layer Name | Type | Output Shape | Parameters | Details |
|---|---|---|---|---|
| **conv2d_2** | Conv2D | (None, 48, 48, 32) | 896 | 32 filters, likely kernel size 3×3 |
| **max_pooling2d_2** | MaxPooling2D | (None, 24, 24, 32) | 0 | 2×2 pool size, reduces spatial dimensions by half |
| **conv2d_3** | Conv2D | (None, 22, 22, 64) | 18,496 | 64 filters, kernel size 3×3, input channels = 32 |
| **max_pooling2d_3** | MaxPooling2D | (None, 11, 11, 64) | 0 | 2×2 pool size |
| **flatten_1** | Flatten | (None, 7744) | 0 | Flattens (11×11×64) output to 1D vector |
| **dense_2** | Dense (F.C.) | (None, 32) | 247,840 | Fully connected layer with 32 neurons |
| **dense_3** | Dense (Output) | (None, 5) | 165 | Output layer with 5 neurons (5-class classification) |

Summary:
**Total Parameters**: 267,397, **Trainable Parameters**: 267,397, Non-**trainable Parameters**: 0.,
**Input Image Size (inferred)**: Likely 50×50×1 or 49×49×1 (based on backtracking conv/pooling layers), **Output Classes**: 5

## IV. Results and Discussion

The plots, ***(as shown in Fig. 8),*** give important information on the dynamics of the training of the CNN, both the loss and accuracy measures over the training epochs. The training loss curve presented in the left hand panel (in red) shows a gradual decrease as the training proceeds. This steady drop means that this model is successfully absorbing the trends of the training data and the loss value is near to a lower limit, which means better predictive ability on the training set. By contrast, the validation loss, indicated in green, starts to exhibit a similar trend to the training loss but it has a downward trend which indicates successful generalisation at the early stages of training. But with the continued training, there is a slight increase in the validation loss, which indicates the start of over-fitting where the model will start memorising the training data and lose its ability to predict on unseen data. The strongly indicated mark with the label ***best epoch (at epoch 10)*** indicates the most ideal time to switch the model before the drop and thus, the conclusion of training should be optimally at this epoch in order to retain the generalisation skills.

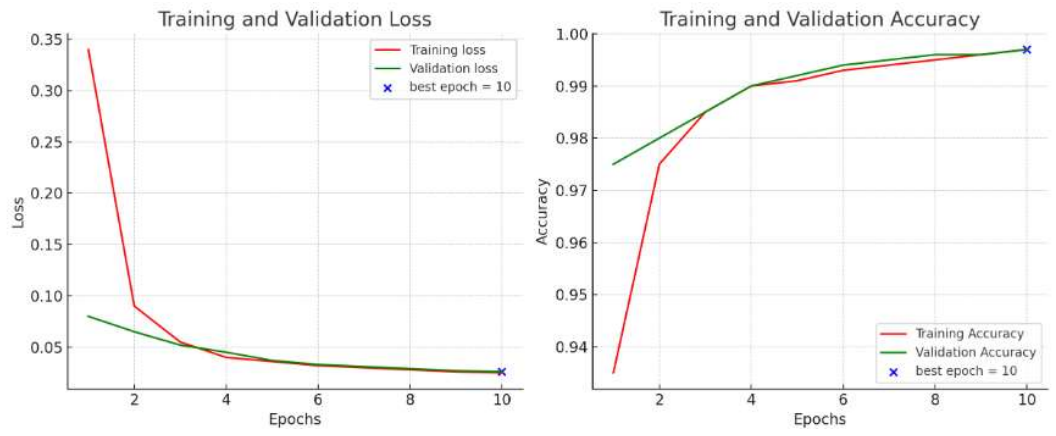

Fig. 8. Training loss and accuracy curves illustrating model performance over epochs.

As illustrated in the ROC curve ***(as shown in Fig. 9)***, convolutional neural network is quite powerful in differentiating between the different varieties of rice, as the network can classify all the classes perfectly. This high performance shows that the model is appropriately tuned to be used in applications that need a good classification of these varieties. The ROC curve is a tool, which cannot be skipped in the visualization of the *trade-off* between sensitivity and specificity to strengthen the reliability of the model in the real-world scenario.

TABLE 3. DETAILED CLASSIFICATION REPORT OF CNN MODEL FOR DIFFERENT RICE GRAIN TYPES

| Parameter Name | Precision | Recall | F1-Score |
|---|---|---|---|
| Arborio | 0.987 | 0.985 | 0.986 |
| Basmati | 0.993 | 0.989 | 0.991 |
| Ipsala | 0.998 | 1.00 | 0.997 |
| Jasmine | 0.986 | 0.991 | 0.989 |
| Karacadag | 0.987 | 0.992 | 0.990 |

The images ***(as shown in Fig. 10)*** inscribed with a label of ***"Original Image"*** depicts a grain of rice and serves as the visual data to the classification model. This kind of representation is necessary to train the model to identify and distinguish the different types of rice with their visual features. The quality and clarity of the input images are the direct determinants of the ability of the model to learn and produce accurate predictions. The images ***(as shown in Fig. 10)***, marked with ***"LIME Explanation"***, provides the outcomes of the LIME (Local Interpretable Model-agnostic Explanations) method on the predictions of the CNN. In this case, the yellow line draws attention to the areas of the rice grain that were found to have the greatest impact in the model in its classification decision. Besides increasing the model interpretability, this visualization provides information about what features are used in the model to make predictions. On explaining what was important in the image, LIME enables researchers and stakeholders to understand the logic of the model, hence creating a sense of trust and confidence in its deliverables. On the whole, these visualizations reflect the two-fold generally of the research aim, to reach high classification accuracy, but with the transparency of the decision-making process of machine-learning models.

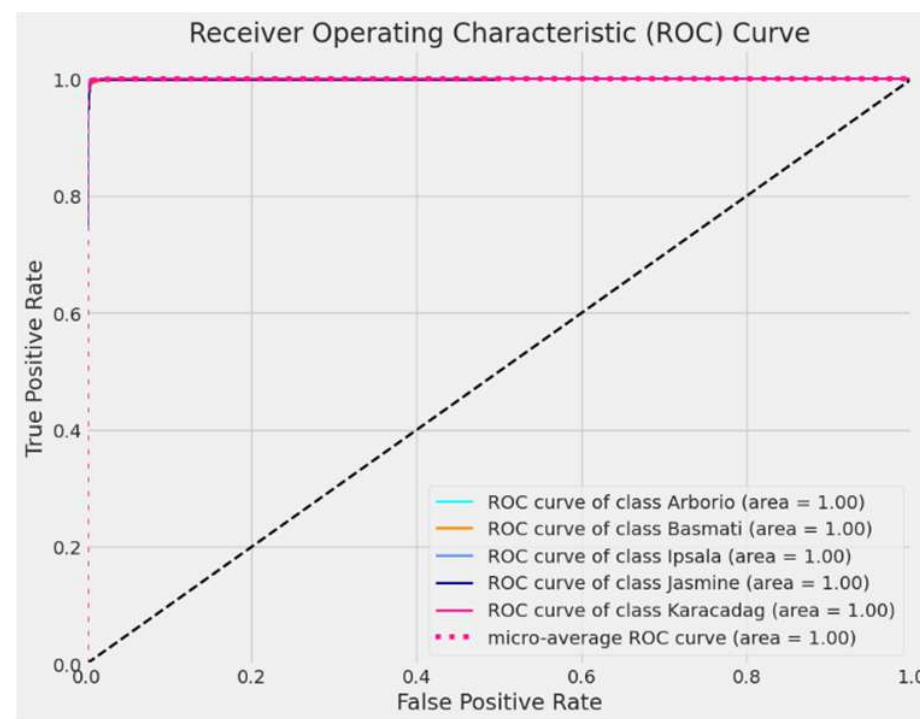

Fig. 9. ROC curve illustrating the trade-off between accurate positive and false positive rates.

The SHAP analysis depicted in the image ***(as shown in Fig. 12)*** provides valuable insights into the factors influencing the classification of Basmati rice grain across multiple outputs in the form of output 0\~3. Each output represents a different class related to the rice variety, and the corresponding SHAP values highlight the contributions of individual pixels to the model predictions.

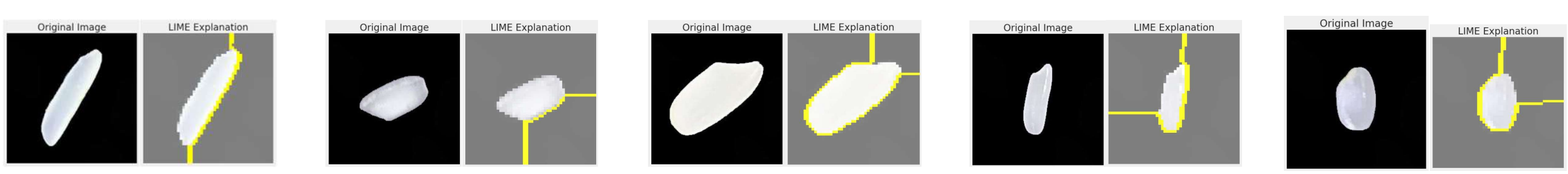

*(a)* *(b)* *(c)* *(d)* *(e)*

Fig. 10. Original Image vs. LIME-generated visualization highlighting influential features (a) Basmati Rice Grain, (b) Arborio Rice Grain, (c) Ipsala Rice Grain, (d) Jasmine Rice Grain, (e) Karacadag Rice Grain

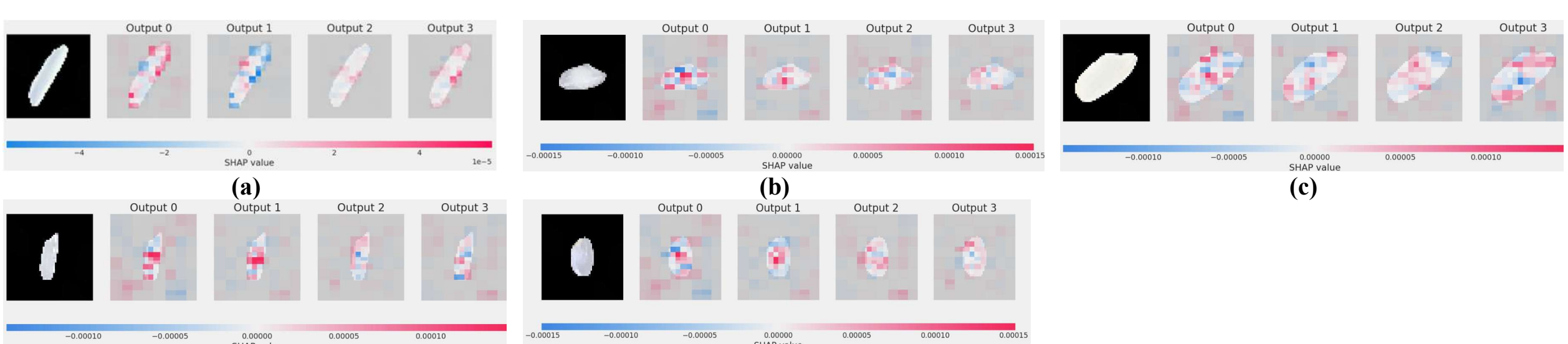

**(a)** **(b)** **(c)** **(d)** **(e)**

Fig. 11. SHAP-generated output highlights the key features: (a) Basmati rice grain, (b) Arborio Rice Grain, (c) Ipsala Rice Grain, (d) Jasmine Rice Grain, (e) Karacadag Rice Grain.

## V. CONCLUSIONS

The current study presents and tests a convolutional neural network (CNN) to classify one of the varieties of rice, such as Arborio, Basmati, Ipsala, Jasmine, and Karacadag. The model performed exceptionally well through tough training and validation achieving high accuracy and a perfect area under ROC curve per class. The confusion matrix revealed that the model had success in differentiating the rice varieties, and its misclassifications were very limited. The model was also found to learn and generalize as the dynamics of the training were analyzed using the loss and accuracy plots. It is important to note that the overfitting to the end of the training is observed, and it is necessary to monitor validation values to maintain model strength. Further, the additions of explainability tools like LIME and SHAP have provided important information on the decision making process of the model. These techniques advance our knowledge on how the model infers the rice grain properties by visualizing the contribution of individual characteristics to classification outcomes, thus developing confidence in predicting rice grain. The above results highlight the promise of deep-learning methods in the field of agriculture with a particular role in enhancing the classification of crops.

## REFERENCES

[1] D. M. K. S. Hemathilake and D. M. C. C. Gunathilake, "Agricultural productivity and food supply to meet increased demands," in *Future foods*, Elsevier, 2022, p. 539–553.

[2] V. Singh, N. Sharma and S. Singh, "A review of imaging techniques for plant disease detection," *Artificial Intelligence in Agriculture,* vol. 4, p. 229–242, 2020.

[3] R. Abiri, N. Rizan, S. K. Balasundram, A. B. Shahbazi and H. Abdul-Hamid, "Application of digital technologies for ensuring agricultural productivity," *Heliyon,* vol. 9, 2023.

[4] O. Rama Devi, M. S. Al Ansari, B. P. Reddy, M. HT, Y. A. Baker El-Ebiary, M. Rengarajan and others, "Optimizing Crop Yield Prediction in Precision Agriculture with Hyperspectral Imaging-Unmixing and Deep Learning.," *International Journal of Advanced Computer Science & Applications,* vol. 14, 2023.

[5] E. Elbasi, N. Mostafa, C. Zaki, Z. AlArnaout, A. E. Topcu and L. Saker, "Optimizing agricultural data analysis techniques through AI-powered decision-making processes," *Applied Sciences,* vol. 14, p. 8018, 2024.

[6] S. K. Swarnkar, L. Dewangan, O. Dewangan, T. M. Prajapati and F. Rabbi, "AI-enabled crop health monitoring and nutrient management in smart agriculture," in *2023 6th international conference on contemporary computing and informatics (IC3I)*, 2023.

[7] R. Banerjee, Bharti, P. Das and S. Khan, "Crop yield prediction using artificial intelligence and remote sensing methods," in *Artificial intelligence and smart agriculture: technology and applications*, Springer, 2024, p. 103–117.

[8] R. Dhanapal, A. AjanRaj, S. Balavinayagapragathish and J. Balaji, "Crop price prediction using supervised machine learning algorithms," in *Journal of physics: Conference series*, 2021.

[9] K. P. Reddy, M. R. Thanka, E. B. Edwin, V. Ebenezer, P. Joy and others, "Farm_{e}{r}{a}: Precision farming with GIS, AI pest management, smart irrigation, data analytics, and optimized crop planning," in *2024 International Conference on Inventive Computation Technologies (ICICT)*, 2024.

[10] A. Uzhinskiy, "Advanced technologies and artificial intelligence in agriculture," *AppliedMath,* vol. 3, p. 799–813, 2023.

[11] S. Yang, L. Gu, X. Li, T. Jiang and R. Ren, "Crop classification method based on optimal feature selection and hybrid CNN-RF networks for multi-temporal remote sensing imagery," *Remote sensing,* vol. 12, p. 3119, 2020.

[12] S. B. Ahmed, S. F. Ali and A. Z. Khan, "On the frontiers of rice grain analysis, classification and quality grading: A review," *IEEE Access,* vol. 9, p. 160779–160796, 2021.

[13] P. Dhiman, A. Kaur, Y. Hamid, E. Alabdulkreem, H. Elmannai and N. Ababneh, "Smart disease detection system for citrus fruits using deep learning with edge computing," *Sustainability,* vol. 15, p. 4576, 2023.

[14] M. M. Hasan, T. Rahman, A. S. Uddin, S. M. Galib, M. R. Akhond, M. J. Uddin and M. A. Hossain, "Enhancing rice crop management: Disease classification using convolutional neural networks and mobile application integration," *Agriculture,* vol. 13, p. 1549, 2023.

[15] S. Tammina and others, "Transfer learning using vgg-16 with deep convolutional neural network for classifying images," *International Journal of Scientific and Research Publications (IJSRP),* vol. 9, p. 143–150, 2019.

[16] C. Szegedy, V. Vanhoucke, S. Ioffe, J. Shlens and Z. Wojna, "Rethinking the inception architecture for computer vision," in *Proceedings of the IEEE conference on computer vision and pattern recognition*, 2016.

[17] A. Howard, M. Sandler, G. Chu, L.-C. Chen, B. Chen, M. Tan, W. Wang, Y. Zhu, R. Pang, V. Vasudevan and others, "Searching for mobilenetv3," in *Proceedings of the IEEE/CVF international conference on computer vision*, 2019.

[18] M. T. Ribeiro, S. Singh and C. Guestrin, "" Why should i trust you?" Explaining the predictions of any classifier," in *Proceedings of the 22nd ACM SIGKDD international conference on knowledge discovery and data mining*, 2016.

[19] S. Rao, S. Mehta, S. Kulkarni, H. Dalvi, N. Katre and M. Narvekar, "A study of LIME and SHAP model explainers for autonomous disease predictions," in *2022 ieee bombay section signature conference (ibssc)*, 2022.

[20] H. Khalid, S. Saqib, M. J. Asif and D. A. Dewi, "Strategic Customer Segmentation: Harnessing Machine Learning For Retaining Satisfied Customers," *Lahore Garrison University Research Journal of Computer Science and Information Technology,* vol. 8, 2024.

[21] M. Abouelyazid, "Advanced artificial intelligence techniques for real-time predictive maintenance in industrial IoT systems: A comprehensive analysis and framework," *Journal of AI-Assisted Scientific Discovery,* vol. 3, p. 271–313, 2023.

[22] M. J. Asif, M. Asad and S. Imran, "Crowd Scene Analysis: Crowd Counting using MCNN based on Self-Supervised training with Attention Mechanism," in *2023 25th International Multitopic Conference (INMIC)*, 2023.

[23] M. J. Asif, "Crowd Scene Analysis Using Deep Learning Techniques," *arXiv preprint arXiv:2505.08834,* 2025.

[24] M. J. Asif, S. Saqib, R. F. Ahmad, M. Asad and S. T. Hussain Rizvi, "Conv-LSTM for Real-Time Spatio-Temporal Analysis of Crowd Behavior in Public Spaces," in *2025 4th International Conference on Communication, Computing and Digital Systems (C-CODE)*, 2025.

[25] B. Abhisheka, S. K. Biswas and B. Purkayastha, "A comprehensive review on breast cancer detection, classification and segmentation using deep learning.," *Archives of Computational Methods in Engineering,* vol. 30, 2023.

[26] Z. Wang, K. Wang, X. Chen, Y. Zheng and X. Wu, "A deep learning approach for inter-patient classification of premature ventricular contraction from electrocardiogram," *Biomedical Signal Processing and Control,* vol. 94, p. 106265, 2024.

[27] F. D. Adhinata, R. Sumiharto and others, "A comprehensive survey on weed and crop classification using machine learning and deep learning," *Artificial intelligence in agriculture,* vol. 13, p. 45–63, 2024.

[28] I. M. Nasir, A. Bibi, J. H. Shah, M. A. Khan, M. Sharif, K. Iqbal, Y. Nam and S. Kadry, "Deep learning-based classification of fruit diseases: An application for precision agriculture," 2021.

[29] M. Koklu, I. Cinar and Y. S. Taspinar, "Classification of rice varieties with deep learning methods," *Computers and electronics in agriculture,* vol. 187, p. 106285, 2021.

[30] A. Khamparia, G. Saini, D. Gupta, A. Khanna, S. Tiwari and V. H. C. de Albuquerque, "Seasonal crops disease prediction and classification using deep convolutional encoder network," *Circuits, Systems, and Signal Processing,* vol. 39, p. 818–836, 2020.

[31] P. Saxena, K. Priya, S. Goel, P. K. Aggarwal, A. Sinha and P. Jain, "Rice varieties classification using machine learning algorithms," *Journal of Pharmaceutical Negative Results,* vol. 13, p. 3762–3772, 2022.

[32] Y. Ramdhani and D. P. Alamsyah, "Enhancing sustainable rice grain quality analysis with efficient svm optimization using genetic algorithm," in *E3S Web of Conferences*, 2023.

[33] G. Sakkarvarthi, G. W. Sathianesan, V. S. Murugan, A. J. Reddy, P. Jayagopal and M. Elsisi, "Detection and classification of tomato crop disease using convolutional neural network," *Electronics,* vol. 11, p. 3618, 2022.

[34] K. M. A. Bejerano, C. C. Hortinela IV and J. J. R. Balbin, "Rice (Oryza Sativa) Grading classification using Hybrid Model Deep Convolutional Neural Networks-Support Vector Machine Classifier," in *2022 IEEE International Conference on Artificial Intelligence in Engineering and Technology (IICAIET)*, 2022.

[35] M. S. Rayudu, L. K. Pampana, S. Myneni and S. Kalavari, "Rice Grain Classification for Agricultural Products Marketing-A Deep Learning Approach," in *2023 First International Conference on Advances in Electrical, Electronics and Computational Intelligence (ICAECI)*, 2023.